\documentclass[journal]{IEEEtran}
\usepackage{amsmath,amssymb,amsfonts,graphicx,booktabs,multirow,array,tabularx,adjustbox}
\usepackage{cite,url,xcolor,hyperref}
\hypersetup{hidelinks}

\newcommand{\Ddev}{\mathcal{D}_{\mathrm{dev}}}

\begin{document}
\title{Can Large Language Models Optimize Pretrained Neural Network Weights? A Multi-Phase Empirical Evaluation of Gradient-Free Editing in YOLO26 Object Detection}
\author{Mohamed Abdallah Salem\\
  Mohamed-salem@ieee.org
  \thanks{}}
\markboth{}{Gradient-Free Weight Editing for YOLO26}
\maketitle
\begin{abstract}
Direct modification of pretrained neural network parameters offers an alternative to gradient-based fine-tuning, but it is unclear whether small, structured changes can consistently improve object detection or whether large language models (LLMs) can guide those changes better than conventional black-box optimizers. This article consolidates five experimental phases involving pretrained YOLO26 detectors and Microsoft COCO. We assess deterministic scaling, random perturbations, differential evolution, covariance matrix adaptation evolution strategy (CMA-ES), selected low-rank and head interventions, genuine local-LLM proposals, feedback- and failure-aware LLM policies, and a hybrid LLM--CMA-ES optimizer with a matched random-policy control. The primary search objective is native Ultralytics mAP50--95, while separately stored COCO API AP50--95 is used for specified validation comparisons. Development search uses 1,000 COCO Val2017 images; a disjoint but previously examined 1,500-image selection set measures transfer of selected edits. The archived multi-budget experiment reports that increasing CMA-ES evaluations from 8 to 250 improved mean development gains from +0.11 to +0.28 percentage points, while the corresponding selection changes remained negative (from -0.03 to -0.12 points); the largest budget has only one seed. Hybrid LLM versus matched random-policy comparisons do not establish superiority ($p=0.612$, $0.369$, and $0.694$ for budgets 8, 25, and 50). A five-seed YOLO26m replication yields an official COCO API AP change of +0.004 percentage points (95\% confidence interval $[-0.026,+0.034]$). A retrospective review of 225 recorded LLM proposals finds 53 that conform to the finalized strict schema, while historical tensor-level attribution remains incomplete for early experiments. The evidence supports the feasibility of genuinely LLM-mediated gradient-free editing and demonstrates development-set selection effects, but does not establish a consistent generalization improvement or an incremental LLM advantage under the tested conditions.
\end{abstract}

\section{Introduction}
\IEEEPARstart{P}{retrained} object detectors are generally adapted through parameter-efficient or full gradient-based fine-tuning. Such procedures require access to a differentiable objective, training software, an optimizer, and often a labeled adaptation set. A different possibility is to intervene directly in existing parameters while leaving the architecture and training procedure unchanged. The resulting checkpoint is evaluated by forward propagation alone. This approach is attractive for controlled scientific interventions, for inspecting parameter sensitivity, and for understanding the limits of low-dimensional black-box adaptation~\cite{hansen2016,storn1997,snoek2012,bergstra2011}.

The broader literature offers precedents for operating in parameter space. Model soups average trained checkpoints, task arithmetic combines task vectors, and model-merging methods reconcile interference between independently trained weights~\cite{wortsman2022,ilharco2023,yadav2023}. Low-rank adaptation and sparse parameter techniques also suggest that meaningful changes can often be represented economically~\cite{hu2022,frankle2019,han2015}. However, these literatures are not direct evidence that an arbitrary edit of an \emph{already pretrained} object detector can outperform the unchanged baseline without learning new gradients. In particular, matching a subset of candidate scores does not guarantee improved detection of unseen images.

Language models provide a further hypothesis. Optimizing by prompting and related search frameworks show that language models can propose new candidates after reading prior objective evaluations~\cite{yang2024opro,shinn2023,yao2023}. Whether language-model-generated explanations translate into useful numerical weight interventions is largely an empirical question. A genuine test requires real inference calls, structured proposal validation, and an ablation in which a nonsemantic policy receives the same numerical search budget. Otherwise, a locally generated Gaussian step can be mislabeled as \emph{LLM-guided optimization}.

Industrial computer vision offers practical examples of learned visual classification outside generic detection benchmarks; for example, Salem \emph{et al.} studied deep-learning-based material recognition for laser cutting~\cite{salem2023laser}. This application illustrates the breadth of deployment settings in which adaptation and reliability warrant evaluation, but it does not constitute a direct weight-editing baseline.\par\noindent We address these questions using pretrained YOLO26 models~\cite{jocher2026,yolov1,redmon2018,bochkovskiy2020} on COCO~\cite{lin2014}. Across five archived phases, initial direct scaling and perturbation lead to stage-wise and evolutionary optimization; the subsequent phases implement genuine LLM proposals, failure diagnostics, hybrid searches, multi-budget comparisons, and model-capacity experiments. Rather than asserting that edited weights improve accuracy, the study tests the extent to which the resulting gains survive on data not used for candidate scoring.

\subsection{Questions and contributions}
The central questions are (RQ1) whether direct editing increases accuracy on an unseen evaluation partition; (RQ2) how random, evolutionary, and LLM methods compare under controlled budgets; (RQ3) which stages and operation families are sensitive; (RQ4) whether growing the search budget helps selection-set accuracy; (RQ5) whether LLM guidance provides an advantage beyond matched random-policy bounding; and (RQ6) whether observed behavior persists from YOLO26n to YOLO26s and YOLO26m.

The contribution is a \emph{systematic evaluation}, not a new trained detector: a bounded weight-intervention implementation; a record of real Ollama-proposed edits and explicit validity checks; matched numerical and nonsemantic controls; multi-budget and cross-capacity results; and an audit that distinguishes retrospective validity from evidence of historical execution. The paper retains negative and non-significant results, labels source-specific metrics explicitly, and separates supported observations from mechanistic hypotheses.

\section{Background and Related Literature}

\subsection{Object detection architectures and benchmarking}
Two-stage detectors such as Faster R-CNN~\cite{ren2015} and single-stage methods such as SSD and RetinaNet~\cite{liu2016,lin2017focal} established complementary detection design spaces. YOLO detectors emphasize real-time prediction and progressively incorporate improved feature fusion, head design, and training procedures~\cite{yolov1,redmon2018,bochkovskiy2020,jocher2026}. Transformer-based approaches, including DETR and deformable DETR, offer additional alternatives for end-to-end detection~\cite{carion2020,zhu2021}. YOLO26 supports end-to-end prediction, and the checkpoint structure used in these experiments contains components with different sensitivities to tensor modification~\cite{jocher2026}. Residual feature extraction and batch normalization are established building blocks whose parameter interactions can affect editing behavior~\cite{he2016,ioffe2015}.

COCO contains 80 detection classes and has standardized object-detection evaluation across multiple intersection-over-union thresholds~\cite{lin2014}. AP depends on confidence ranking, matching, category mapping, box geometry, and the implementation of interpolation. Accordingly, a comparison between checkpoints is interpretable only when evaluated with identical image partitions and inference settings. Software-implemented metrics can differ from official COCO API results even when their names sound alike.

\subsection{Direct parameter interventions and model merging}
Weight averaging, task arithmetic, and merging exploit structure in model parameter spaces~\cite{wortsman2022,ilharco2023,yadav2023,ainsworth2023}. Mode connectivity and permutation symmetries caution against simplistic interpretations of Euclidean parameter distance~\cite{garipov2018,draxler2018,entezari2022}. A coefficient near unity is not universally harmless: scaling convolutional weights can alter normalization behavior, head calibration, and feature magnitudes. Likewise, local robustness is not proof that a model lies at a mathematically sharp optimum; sharpness depends on perturbation parameterization and geometry~\cite{keskar2017,dinh2017}.

Robustness also depends on whether an intervention preserves task-relevant structure. In laser-speckle material classification, Salem and Diab investigated how structural preservation affects the behavior of data augmentation in deep learning~\cite{salemdiab2026speckle}. Although image-space augmentation is distinct from parameter-space editing, this study provides an application-level motivation for explicitly examining which transformations preserve discriminative information rather than assuming that larger perturbations are beneficial. We do not treat the speckle-classification findings as evidence of an object-detection accuracy gain from weight editing.

Low-rank methods provide a compact update parameterization~\cite{hu2022}; network pruning studies explore removal or zeroing of parameters~\cite{han2015,frankle2019}; and knowledge distillation targets different efficiency--accuracy trade-offs~\cite{hinton2015}. Conventional gradient-based optimizers, including Adam and decoupled-weight-decay variants, instead update parameters from training gradients~\cite{kingma2015,loshchilov2019}. The present study does not conflate direct weight editing with these training-based methods. It examines transformations of a fixed checkpoint and records whether accuracy changes under those edits.

\subsection{Numerical gradient-free search}
Differential evolution (DE) performs population-based mutation, recombination and selection~\cite{storn1997}; CMA-ES adapts a search distribution to local objective responses~\cite{hansen2016}; Bayesian optimization and tree-structured Parzen estimators are common hyperparameter-search strategies~\cite{snoek2012,bergstra2011}; and random search is an essential budget-controlled baseline~\cite{bergstra2012}. These algorithms optimize observed metrics through repeated evaluation. Because observations arise from finite samples, adaptive search can select candidates that exploit sample-specific fluctuations. Reusing a validation set across successive phases compounds this risk~\cite{dwork2015,recht2019}.

\subsection{Language models as optimization policies}
LLM-based optimization may be formulated as iterative generation of candidate decisions conditioned on prior responses~\cite{yang2024opro,shinn2023,yao2023}. The generated explanation is not itself a validation of the candidate's effect. In this study the LLM selects stages or structured operations, while an external evaluator supplies actual COCO scores. Unlike gradient-based fine-tuning, the LLM does not compute or update a loss gradient. Unlike a simulated textual optimizer, genuine use requires an inference endpoint, saved request and response metadata, and explicit validation of the proposed patch.

\section{Study Design and Experimental Resources}
\subsection{Five-phase progression}
Phase I establishes original YOLO26 baseline measurements, deterministic stage scaling, stochastic perturbation, and sensitivity profiling. Phase II develops numerical comparison methods, recorded real-LLM optimization, structural low-rank and head edits, and paired COCO bootstrap evaluation. Phase III adds failure diagnostics and hybrid LLM--CMA-ES policies. Phase IV introduces strict action validation, a historical LLM audit, a random-policy hybrid control, and five optimization seeds at short budgets. Phase V extends budget values to 8, 25, 50, 100 and 250 evaluations, evaluates multiple model capacities and replicates an exploratory YOLO26m observation. The main results presented below are \emph{archived measurements}; no new model inference was executed during manuscript preparation.

\subsection{Hardware, software, and checkpoint characteristics}
The source report lists two NVIDIA RTX 4500 Ada accelerators with 24 GB VRAM per GPU, Linux, Python 3.13.7, PyTorch 2.8.0+cu128, CUDA 12.8, and Ultralytics 8.4.16. The experiments use YOLO26n/s/m checkpoint files. Fused parameter counts differ from unfused or training-head parameter counts; the manuscript uses the final report's corresponding fused counts where possible. Numerical results should not be compared to vendor benchmark results without matching all inference settings~\cite{jocher2026,paszke2019}.

\begin{table}[t]
\caption{Archived detector baselines on COCO Val2017 subsets. All values are percentages except parameters. Official COCO API and native Ultralytics metrics are shown separately.}
\label{tab:baselines}
\centering\scriptsize
\begin{tabular}{@{}llrrr@{}}
\toprule
Model & Subset & Params (M) & COCO AP & Native mAP\\
\midrule
YOLO26n & Dev (1,000) & 2.409 & 42.19 & 42.08\\
         & Selection (1,500) & & 39.25 & 39.17\\
YOLO26s & Dev (1,000) & 9.496 & 48.98 & 48.87\\
         & Selection (1,500) & & 46.57 & 46.44\\
YOLO26m & Dev (1,000) & 20.411 & 54.33 & 54.17\\
         & Selection (1,500) & & 51.31 & 51.17\\
\bottomrule
\end{tabular}
\end{table}

\subsection{Dataset partitioning and provenance}
The core experiment begins with COCO Val2017 (5,000 images). Development contains 1,000 images; the selection partition contains 1,500 disjoint images; a further 2,500-image holdout was evaluated in Phase II. Importantly, because the selection and holdout partitions have been inspected in prior phases, they cannot subsequently be described as untouched confirmatory tests. Stratified selection is reported in the project archive, but image-level membership should be reconstructed from its manifests for independent replication.

The history of the additionally named \emph{fresh} 1,000-image subset is inconsistent. The Phase III execution report explicitly describes downloading it from COCO Train2017, while the final Phase V narrative describes it as drawn exclusively from Val2017. No reconciled image-ID provenance was available in the supplied package to resolve that contradiction. Because pretrained YOLO26 checkpoints are COCO-trained, we neither regard this subset as a new, independent test set nor use its score to support the main generalization conclusion. It is reported only as an ancillary archived measurement when needed. This disclosure avoids overstatement and preserves the distinction between a new subset for WSP and images unseen during original checkpoint pretraining.

\begin{figure*}[t]
\centering
\includegraphics[width=.98\textwidth]{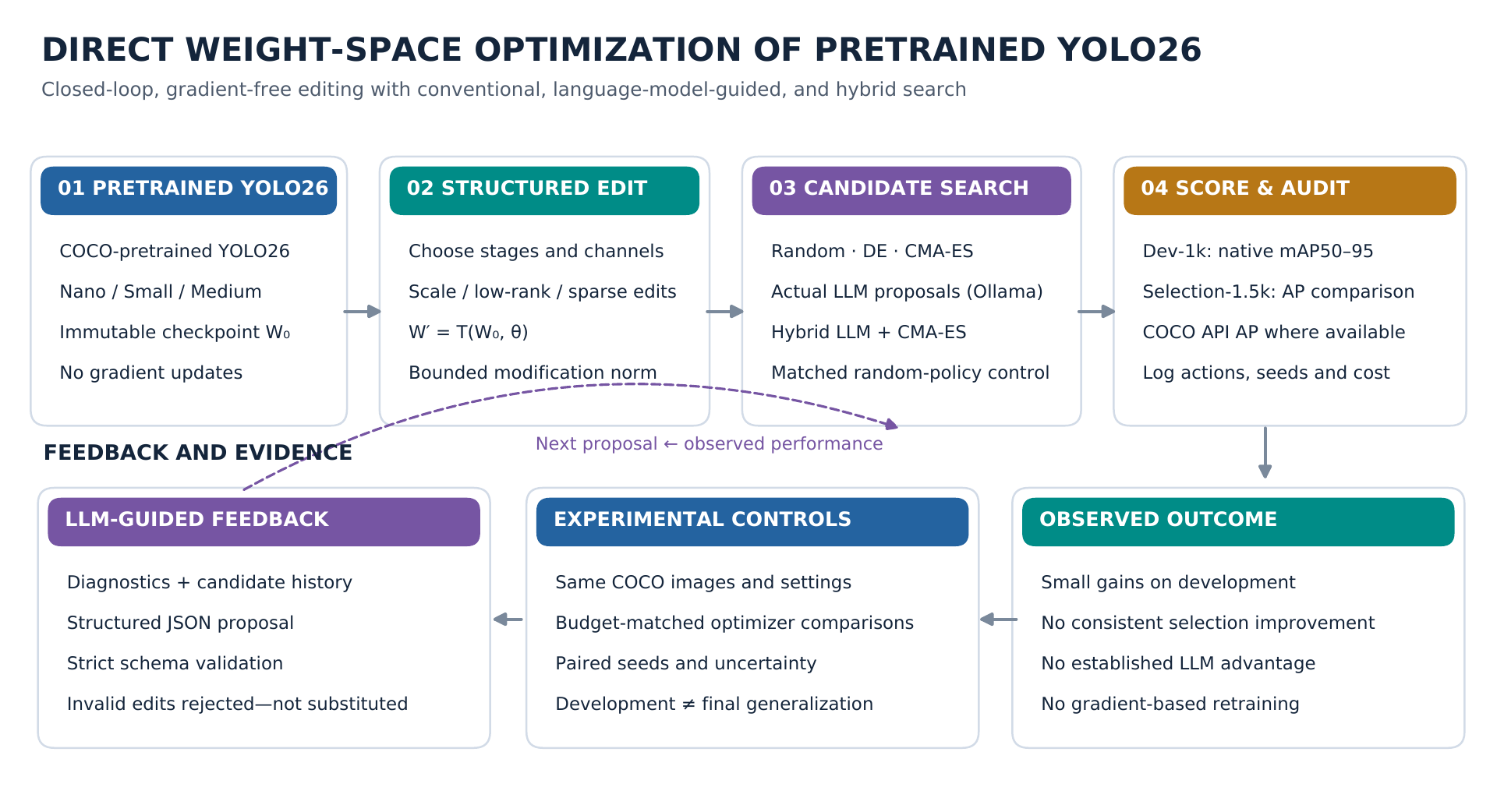}
\caption{Direct Weight-Space Programming framework for pretrained YOLO26. A frozen checkpoint is modified through bounded, gradient-free edits proposed by deterministic, random, evolutionary, or genuine LLM-guided strategies. Candidate performance is measured on development images, while selected outcomes are separately assessed using COCO API evaluation. Observed development gains do not establish out-of-sample improvements.}
\label{fig:workflow}
\end{figure*}

\section{Gradient-Free Weight-Space Methodology}
\subsection{Problem definition and information budget}
Let $f(x;W_0)$ be a pretrained detector, $T$ a validated edit operator, and $\boldsymbol\theta$ the compact continuous or structured edit specification. Define
\begin{equation}
 W_{\boldsymbol\theta}=T(W_0,\boldsymbol\theta),\qquad \boldsymbol\theta\in\Theta.
\label{eq:operator}
\end{equation}
The primary objective in the saved search code is
\begin{equation}
\boldsymbol\theta^*=\arg\max_{\boldsymbol\theta\in\Theta} m_{\mathrm{native}}(W_{\boldsymbol\theta};\Ddev),
\label{eq:search}
\end{equation}
where $m_{\mathrm{native}}$ is the recorded Ultralytics mAP50--95, not necessarily identical to official COCO API AP. No labels or candidate scores from a pristine test set should be supplied during search. Candidate checkpoints are reconstructed from the unchanged original checkpoint to prevent unintended accumulation of successive changes.

\subsection{Scalar, structured, and stochastic edits}
A block-scaling edit is defined by
\begin{equation}
 W'_j=(1+\theta_j)W_{0,j},\quad W'_i=W_{0,i}\;(i\neq j).
\label{eq:block}
\end{equation}
The random perturbation intervention is
\begin{equation}
 W'_j=W_{0,j}+\lambda \epsilon_j,\quad \epsilon_j\sim\mathcal{N}(0,\sigma_j^2).
\end{equation}
Channel-group edits apply selected coefficients to subsets of channels; detection-head edits target regression or classification branches; and low-rank edits implement
\begin{equation}
 W'_j=W_{0,j}+\lambda A_jB_j,\qquad \operatorname{rank}(A_jB_j)\leq r.
\end{equation}
Two compatible modified states may be interpolated as $W_{\beta}=(1-\beta)W_0+\beta W_a$, where $W_a$ is an archived candidate. Magnitude-based sparsification zeroes selected weights. These operations are related to, but distinct from, model merging and trained low-rank adaptation~\cite{wortsman2022,hu2022,yadav2023}.

A bounded relative perturbation is characterized by
\begin{equation}
\rho(W',W_0)=\frac{\|W'-W_0\|_F}{\|W_0\|_F+\varepsilon}, \qquad \rho\leq\tau.
\label{eq:norm}
\end{equation}
The source study uses conservative edit constraints, including a 1\% nominal relative norm ceiling for expanded actions. The actual magnitudes observed in a retrospective audit are much smaller; the ceiling is a \emph{constraint}, not the size of each executed modification.

\subsection{Sensitivity screening and editing locality}
Phase I screened stage-wise small scaling interventions and observed heterogeneous responses. The final detection head was reported as substantially more sensitive than some transition convolutions; specific stage indices and sensitivity values are meaningful only for the verified checkpoint layout. Later head-specific editing showed different behavior between classification and regression branches. A small classification scale shift changed recall and AP; regression-head changes generally reduced AP more severely. These observations motivate localized rather than network-wide editing, but they do not prove causality at the feature level.

\begin{figure*}[t]
\centering\includegraphics[width=.73\textwidth]{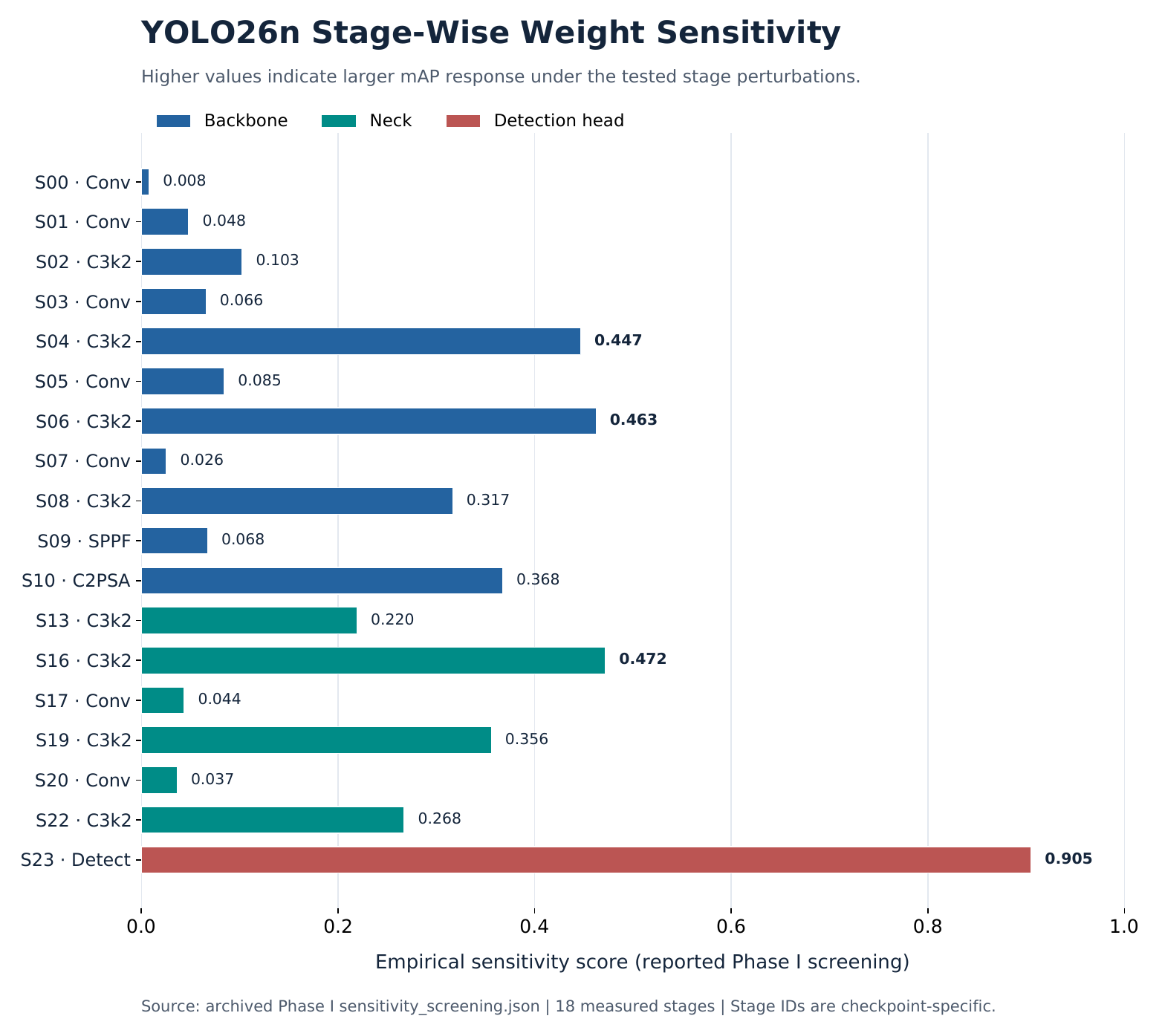}
\caption{Stage-wise sensitivity of pretrained YOLO26n to direct weight modifications, reconstructed from archived Phase I screening measurements. Colored bars distinguish backbone, neck, and detection-head stages. Sensitivity values describe the tested perturbations and checkpoint only; they are not universal layer properties.}
\label{fig:sens}
\end{figure*}

\subsection{Evolutionary and heuristic search}
In random search, candidate coefficients are sampled from prescribed bounds. In DE, population differences generate new proposals~\cite{storn1997}. In CMA-ES, the mean and covariance of a sampling distribution are adapted from candidate objective evaluations~\cite{hansen2016}. The numerical methods receive only scalar development scores. The source package additionally contains Bayesian optimization and coordinate-search implementations; these are supporting experiments rather than uniformly represented methods across every budget.

The optimizer budget $B$ denotes candidate evaluations including any no-change baseline evaluation recorded by a method. The effective number of newly modified candidates may therefore be $B-1$. In the Phase V archive, replicate counts differ by budget: five seeds at $B=8$ and $25$ for major numerical methods, three at $B=50$, two at $B=100$, and one at $B=250$. Thus the larger-budget rows should be interpreted as exploratory optimization traces, not precision estimates of average performance.

\subsection{Genuine LLM proposal generation}
The genuine LLM interface sends structured prompts to a local Ollama model identified by the logs as \texttt{gemma4:latest}. Context can include stage identifiers and weight statistics; evaluation history; permitted edit types and parameter bounds; and, for failure-aware operation, false-positive and false-negative summaries. A proposed JSON action is parsed and checked for field names, permissible target layers, operation types, finite coefficients and edit-magnitude constraints before any weight update.

One practical change across phases is particularly important. An earlier YOLO26--COCO class with the label \texttt{LLMGuidedOptimizer} merely sampled Gaussian perturbations using NumPy. That version is a \emph{heuristic local search}, not actual LLM inference. Later phases retained it as a control and implemented real requests to Ollama. All claims about LLM behavior here refer to recorded genuine responses, not to the original simulator.

\subsection{Failure-aware and hybrid actions}
The failure diagnostic interface calculates operational summaries of background false alarms, localization mismatches, class confusions, confidence distributions and object-size recall. These summaries depend on the archive's own matching implementation; they are auxiliary prompts, not official COCO evaluation metrics. The failure-aware LLM can choose an allowed operation and target; the hybrid method constrains an LLM-specified stage subspace before numerical CMA-ES optimization. The matched random-policy hybrid samples a comparable set of stages and optimization bounds without reading semantic diagnostic context.

\begin{figure*}[t]
\centering\includegraphics[width=.95\textwidth]{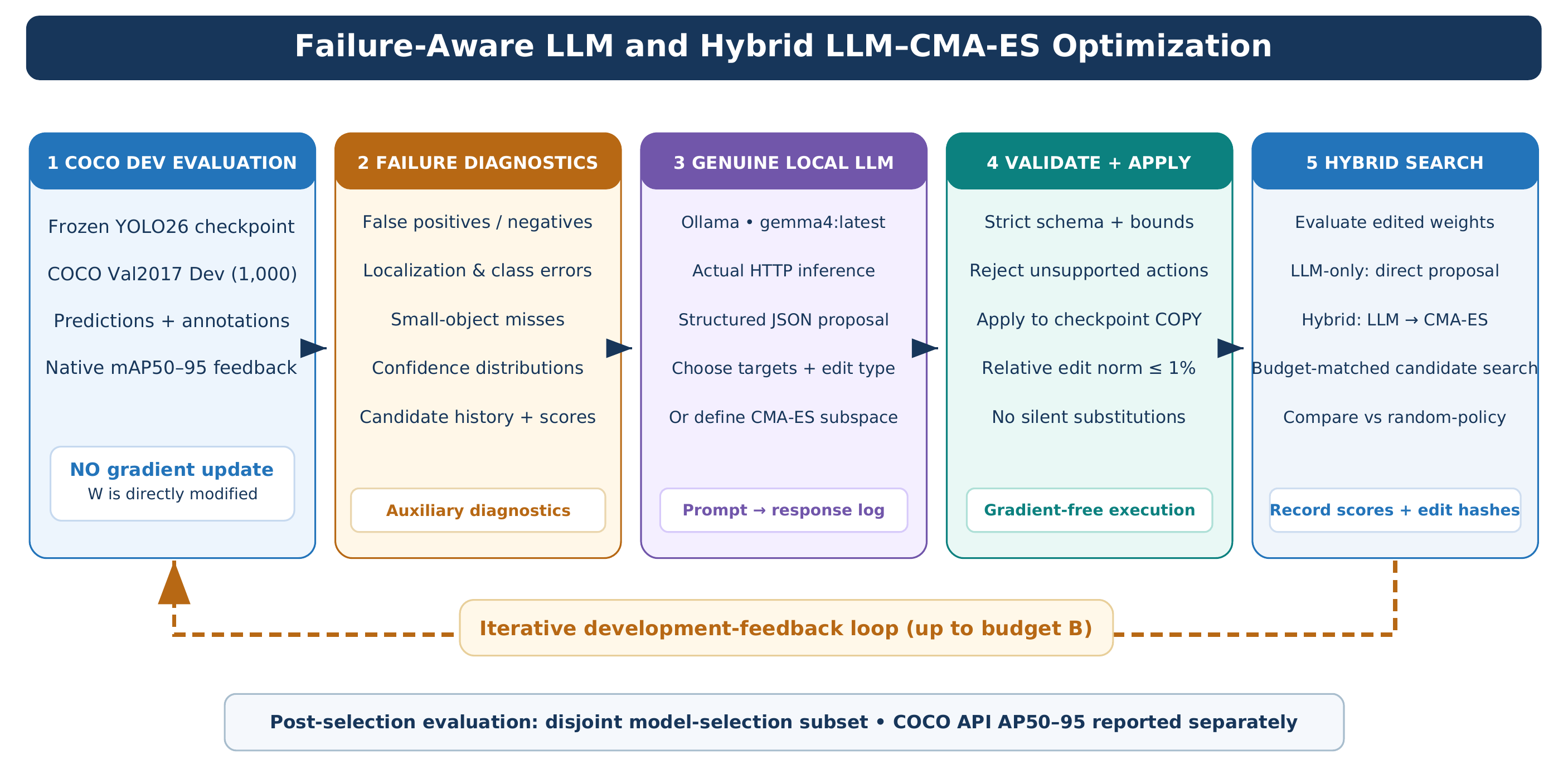}
\caption{Failure-aware LLM and hybrid LLM--CMA-ES optimization architecture. Development-set detections and error diagnostics are summarized for genuine local LLM inference; proposed actions are schema-validated before gradient-free edits, with CMA-ES optimizing bounded subspaces in the hybrid variant. Candidate scores feed the development loop; model-selection evaluation is separate.}
\label{fig:failure}
\end{figure*}

\subsection{Action validity and historical attribution}
A proposal is not identical to a successful edit. The six-level audit distinguishes HTTP responses, extracted JSON, contemporaneous validation, actual runtime tensor modifications, retrospective strict validation and retrospective reconstruction. The source archive contains 225 historical proposal records. Under the \emph{final} strict schema, 53 are accepted and 172 rejected. This is not the original online failure rate: earlier runs used different schemas. Original intermediate model hashes are missing for some early proposals, and the historical code included fallback branches. Consequently, neither successful LLM response nor retrospective reconstructibility proves that the original run executed exactly the intended edit.

\section{Evaluation Protocol and Statistical Analysis}
\subsection{Primary metrics and paired comparisons}
COCO API AP averages precision across IoU thresholds 0.50:0.05:0.95 and the evaluation's recall sampling convention~\cite{lin2014}. The development objective stored by \texttt{CandidateEvaluator} is native Ultralytics mAP50--95. We keep these quantities separate. Percentage-point differences are computed as
\begin{equation}
\Delta_{s}(W')=100\left[m(W';\mathcal{D}_s)-m(W_0;\mathcal{D}_s)\right],
\end{equation}
where $s$ identifies development, selection, or another explicitly named partition; $m$ must be the same software metric for both terms.

For a seed-paired comparison between LLM and random-policy hybrids, let $d_i=\Delta_{\rm LLM,i}-\Delta_{\rm rand,i}$. The reported paired statistic is
\begin{equation}
 t=\frac{\bar d}{s_d/\sqrt{n}},\qquad d_z=\frac{\bar d}{s_d},
\label{eq:paired}
\end{equation}
with $n-1$ degrees of freedom. A nominal two-sided 95\% interval is $\bar d\pm t_{0.975,n-1}s_d/\sqrt n$. These tests assess detectable mean differences under strong assumptions with only 3--5 pairs; they do \emph{not} establish equivalence. A TOST-style equivalence conclusion would require a predeclared effect margin and sufficient power~\cite{lakens2017}.

\subsection{Bootstrapped AP uncertainty}
A Phase II validation computes image-level paired bootstrap replicates with replacement, reconstructing duplicate image identities to avoid incorrect COCO matching. Since AP is a dataset-level statistic, naive independent-object resampling would not be justified. Bootstrap inference is sensitive to sample construction and multiple candidate selection; a confidence interval from one selected checkpoint should not be interpreted as a proof about the full search algorithm~\cite{efron1993,efron1979}.

\subsection{Evaluation-set reuse}
Development images are evaluated repeatedly by construction. Selection images are disjoint but were observed in earlier phases, and the Phase II holdout was also examined before the final protocol. This means conclusions are about archived partition behavior, not a fully prospective preregistered test. Adaptive reuse of validation data can lead to optimistic estimates even if no backpropagation is performed~\cite{dwork2015,recht2019}. Accordingly, the most robust negative finding is \emph{lack of demonstrated consistent improvement}, rather than a claim of inability to improve on any future dataset.

\section{Results I: Baselines, Deterministic Edits, and Structural Sensitivity}
\subsection{Original checkpoint performance}
Table~\ref{tab:baselines} reports exact baseline metric pairs from the final scientific report. Dev AP is higher than selection AP for each model, as expected for two finite, non-identical samples; this difference should not be attributed to editing. The model size comparison also cannot by itself establish differential editability, because architecture and checkpoint weights differ simultaneously.

\subsection{Phase I scaling and random perturbation}
Under the initial development metric, original YOLO26n scored approximately 42.08\% native mAP50--95. Global scaling $\alpha=0.995$ scored 41.94\%, and $\alpha=1.005$ scored 41.71\%; larger global scales such as 0.95 or 1.05 substantially degraded detection. The Phase I deterministic summary's best stage-8 case was near baseline in mAP (42.02\%) but increased reported recall from 51.62\% to 53.45\%. This is a metric trade-off, not a general detector improvement. In random Gaussian perturbations, one isolated trial at $\lambda=0.002$ reached 42.12\% on development, about +0.04--0.05 points, while average performance at higher noise generally declined. These pilot values motivate bounded search and should not be compared directly with final holdout AP.

\begin{table}[t]
\caption{Representative Phase I dev-only interventions on YOLO26n (native mAP50--95, rounded). Negative \(\Delta\) is relative to the source study's approximately 42.08\% baseline.}
\label{tab:phase1}
\centering\footnotesize
\begin{tabular}{lrr}
\toprule
Intervention & Dev mAP (\%) & $\Delta$ (pp)\\
\midrule
Unchanged baseline & 42.08 & 0\\
Global scale $0.995$ & 41.94 & -0.14\\
Global scale $1.005$ & 41.71 & -0.37\\
Global scale $0.95$ & 25.87 & -16.21\\
Stage 8 scale $0.99$ & 42.02 & -0.06\\
Best isolated Gaussian trial & 42.12 & +0.04\\
\bottomrule
\end{tabular}
\end{table}

\subsection{Low-rank, head, sparsity and interpolation interventions}
The Phase II expanded-editing matrix contains multiple ranks and bases, separate classification and regression head perturbations, sparsification from 1--20\%, and interpolation between the base model and a chosen edited model. Illustrative dev-only outcomes are given in Table~\ref{tab:structural}. Some low-rank and classification-head interventions exceeded baseline by less than 0.1 points, while stronger regression-head perturbations were harmful. An interpolation coefficient of 0.8 achieved an exploratory +0.086 points, but without independent confirmation. One percent magnitude pruning was nearly neutral on this metric and does not automatically imply improved hardware latency.

\begin{table}[t]
\caption{Selected archived Phase II structural-edit outcomes (native dev mAP50--95). These are \emph{exploratory}, not independently validated improvements.}
\label{tab:structural}
\centering\footnotesize
\begin{tabular}{p{.61\columnwidth}r}
\toprule
Intervention & $\Delta$ (pp)\\
\midrule
Stage 4, rank 2, residual SVD & +0.052\\
Stage 4, rank 2, random basis & +0.062\\
Classification head, $\alpha=1.005$ & +0.095\\
Regression head, $\alpha=1.015$ & -1.203\\
Magnitude pruning, 1\% & +0.008\\
Magnitude pruning, 20\% & -0.543\\
Base/champion interpolation, $\beta=0.8$ & +0.086\\
\bottomrule
\end{tabular}
\end{table}

\begin{figure}[t]
\centering\includegraphics[width=\columnwidth]{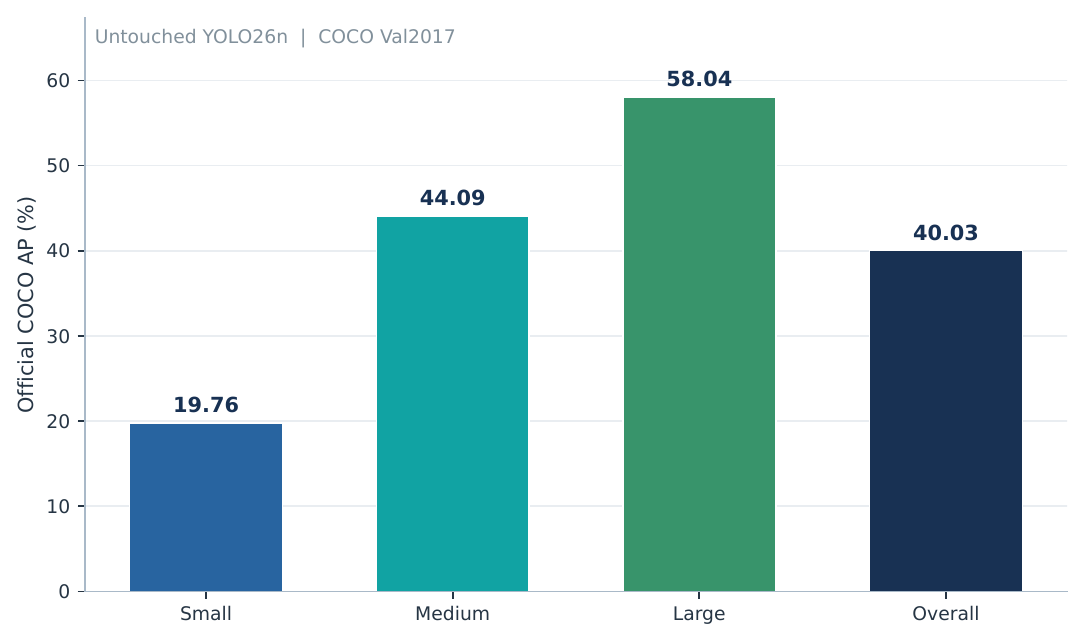}
\caption{Untouched YOLO26n detection AP by object size on full COCO Val2017 (official COCO API). The original archived visualization added illustrative champion gains that were not derived from measured predictions; those unsupported values have been removed.}
\label{fig:size}
\end{figure}

\begin{figure*}[t]
\centering\includegraphics[width=.89\textwidth]{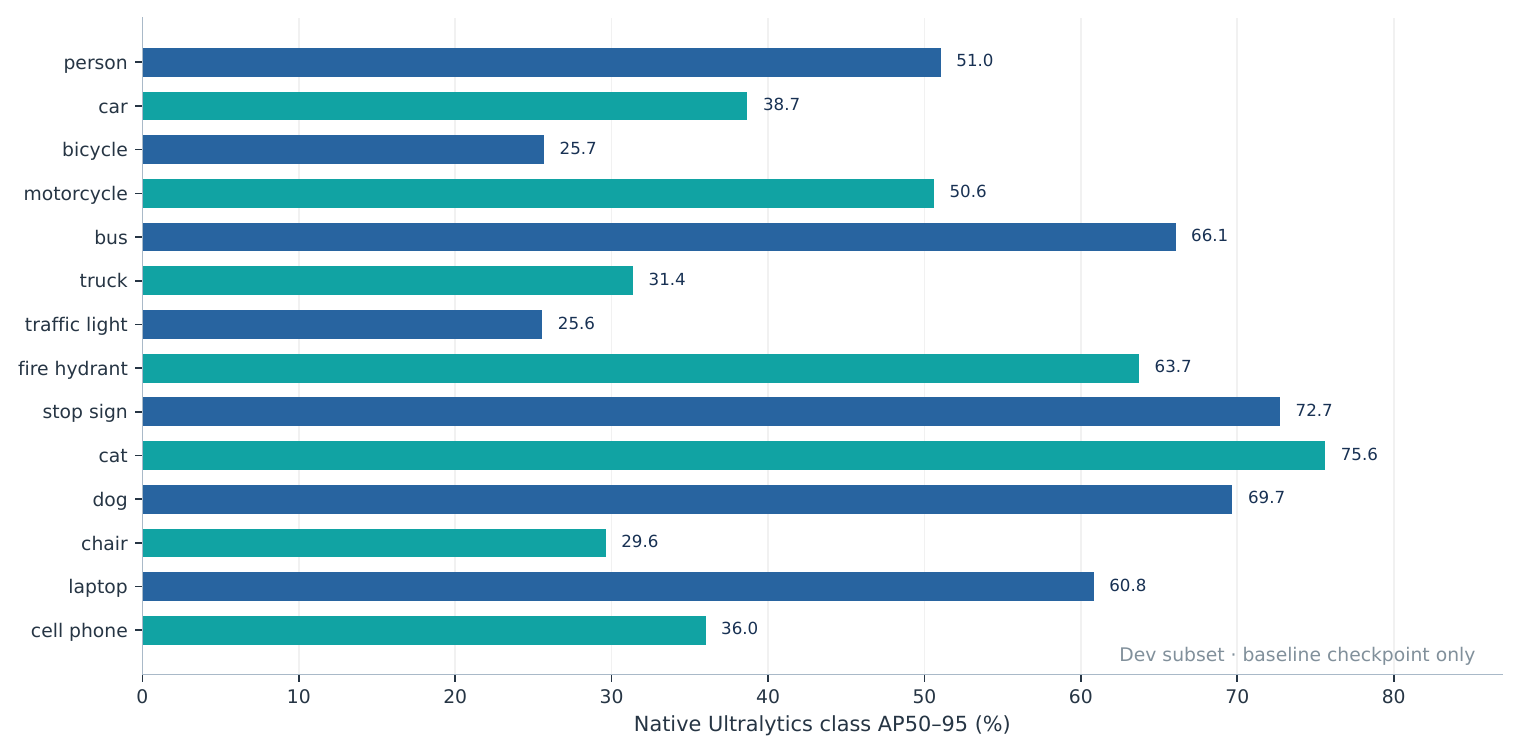}
\caption{Selected per-class native AP50--95 values for the untouched YOLO26n development-set baseline. The earlier plot generated synthetic champion offsets and is replaced by this source-backed baseline diagnostic; no class-wise effect of editing is inferred.}
\label{fig:class}
\end{figure*}

\section{Results II: Numerical Search and Held-Out Checks}
\subsection{Phase II algorithms and bootstrap result}
Phase II included random, DE, CMA-ES and Bayesian-style searches. A selected TPE candidate from its five-dimensional space reached a reported official COCO API AP of 40.9668\% versus 40.9088\% for the YOLO26n baseline on the 2,500-image Phase II holdout; the difference was +0.058 percentage points. Its paired bootstrap interval was approximately $[-0.026,+0.123]$ percentage points, with the archived two-sided $p=0.124$. The result is compatible with a small improvement and a small degradation, and does not establish a reliable benefit. The selected candidate did not originate from the real LLM method.

\begin{table}[t]
\caption{Phase II archived selected-champion evaluation on the 2,500-image holdout (official COCO API AP). This holdout was subsequently examined and is no longer pristine for later method development.}
\label{tab:bootstrap}
\centering\footnotesize
\begin{tabular}{lr}
\toprule
Metric & Value\\
\midrule
Unchanged YOLO26n AP & 40.9088\%\\
TPE-selected WSP AP & 40.9668\%\\
Observed paired change & +0.0580 pp\\
Bootstrap 95\% interval & [-0.026, +0.123] pp\\
Archived two-sided $p$ & 0.124\\
\bottomrule
\end{tabular}
\end{table}

\subsection{Phase III seven-method comparison}
The Phase III method matrix used a 15-evaluation budget per approach, including baseline evaluations where applicable. On development, DE and CMA-ES reached approximately +0.13 and +0.12 points, while real LLM methods produced smaller positive changes (approximately +0.03 to +0.06 points). The hybrid improved the development objective by about +0.01 points. Their subsequent selection and ancillary ``fresh'' measurements did not support a consistent LLM improvement, and the latter subset cannot be interpreted as genuinely unseen until provenance is reconciled. A single short-budget comparison cannot determine the full efficiency ranking of these algorithms.

\begin{figure*}[t]
\centering\includegraphics[width=.89\textwidth]{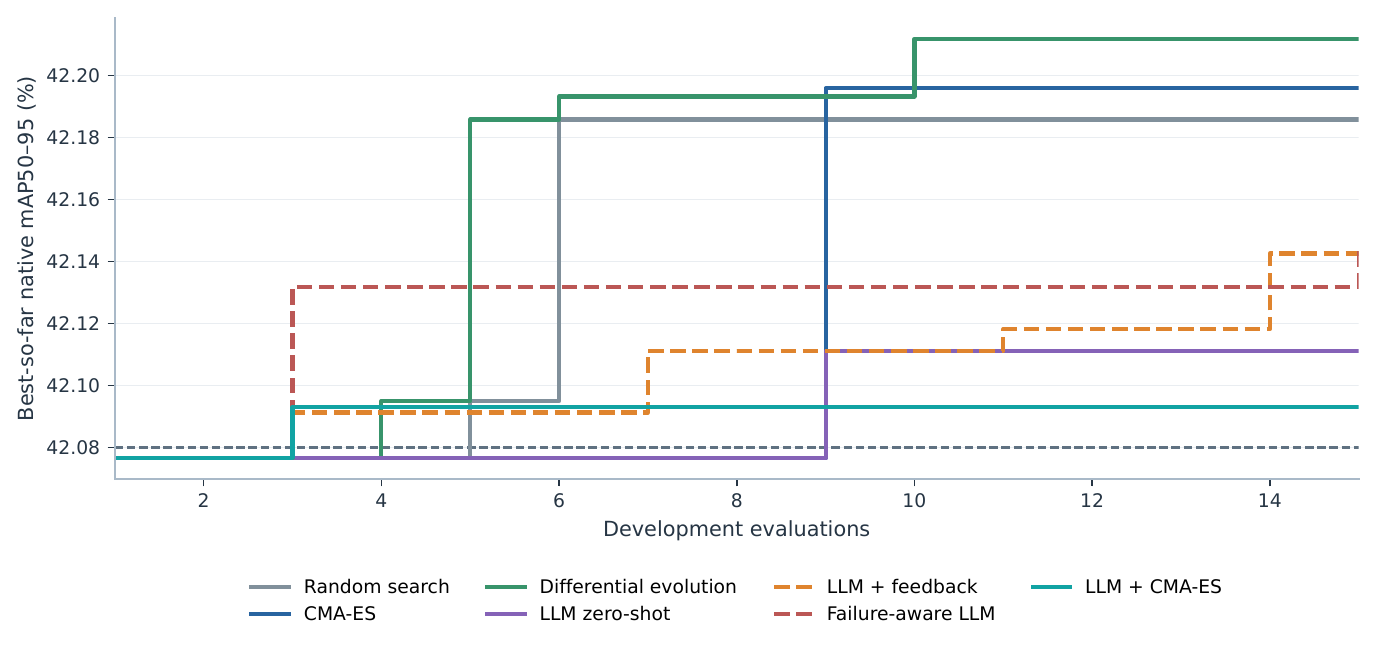}
\caption{Best-so-far native development mAP50--95 over 15 candidate evaluations in the archived Phase III benchmark. Each trace is reconstructed from saved candidate scores. Differences reflect a short exploratory search and do not establish generalization.}
\label{fig:traject}
\end{figure*}

\subsection{Phase IV multi-seed results}
Phase IV increased the number of seeds to five for the seven-method eight-evaluation experiment, adding a matched random-policy hybrid. Most numerical strategies produced larger development-set increases than direct LLM proposals but selection changes near or below zero. The random-policy control was essential: if both hybrids benefit from bounded subspaces and CMA-ES, any additional benefit of the LLM must be measured against that control rather than against an unedited detector alone.

\begin{figure*}[t]
\centering\includegraphics[width=.89\textwidth]{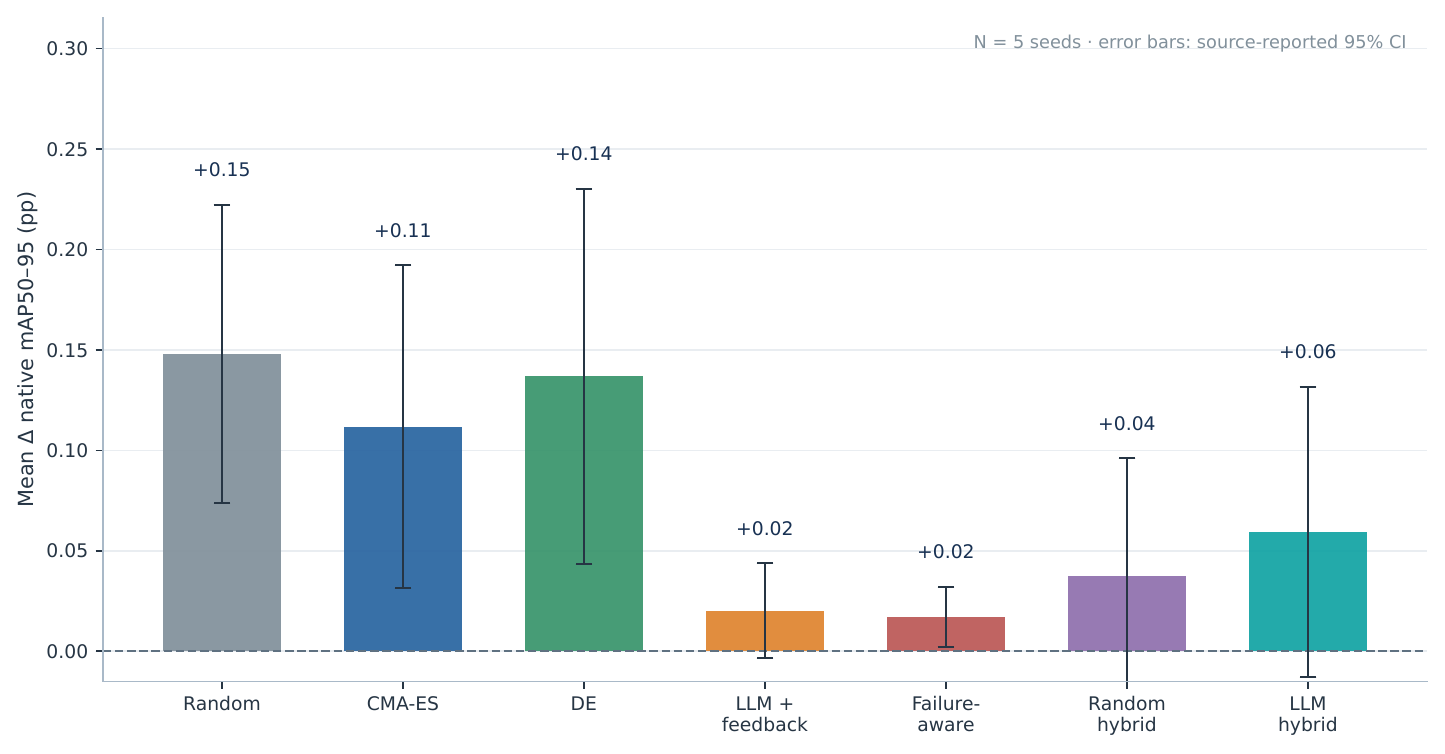}
\caption{Mean development mAP50--95 change across five Phase IV seeds ($B=8$). Error bars show the source-reported 95\% confidence intervals. These are short-budget development results, not independently established gains.}
\label{fig:seeds}
\end{figure*}

\section{Results III: Multi-Budget Phase V Comparisons}
\subsection{Primary archived numerical matrix}
Table~\ref{tab:budgets} reproduces the principal numerical trends from the \emph{packaged Phase V Table 17}, with budget-specific seed counts. The values in the supplied revised prose report differ from the packaged raw-table series for some rows. We use this traceable packaged table here and do not silently average or merge the incompatible versions. All numbers below are native Ultralytics development and selection percentage-point changes, not official COCO API AP.

\begin{table*}[t]
\caption{Phase V numerical optimization budget scaling on YOLO26n (source: archived \texttt{table\_17\_multibudget\_optimization.md}). Values are \(\Delta\) native Ultralytics mAP50--95, in percentage points. Replication count is budget dependent; \(N=1\) is descriptive only.}
\label{tab:budgets}
\centering\small
\begin{tabular}{rrlrrrr}
\toprule
$B$ & Seeds & Optimizer & Dev mean $\Delta$ & Dev SD & Selection mean $\Delta$ & Mean time (s)\\
\midrule
8 & 5 & Random search & +0.15 & 0.06 & -0.11 & 29.1\\
8 & 5 & CMA-ES & +0.11 & 0.06 & -0.03 & 35.9\\
8 & 5 & Differential evolution & +0.14 & 0.08 & -0.05 & 20.7\\
25 & 5 & Random search & +0.19 & 0.03 & -0.07 & 72.5\\
25 & 5 & CMA-ES & +0.17 & 0.03 & -0.06 & 46.1\\
25 & 5 & Differential evolution & +0.18 & 0.04 & -0.08 & 35.6\\
50 & 3 & Random search & +0.20 & 0.03 & -0.07 & 68.8\\
50 & 3 & CMA-ES & +0.20 & 0.03 & -0.06 & 69.2\\
50 & 3 & Differential evolution & +0.19 & 0.03 & -0.12 & 95.8\\
100 & 2 & Random search & +0.21 & 0.02 & -0.11 & 137.8\\
100 & 2 & CMA-ES & +0.25 & $\approx 0$ & -0.05 & 138.2\\
250 & 1 & Random search & +0.23 & N/A & -0.13 & 414.7\\
250 & 1 & CMA-ES & +0.28 & N/A & -0.12 & 414.2\\
\bottomrule
\end{tabular}
\end{table*}

The numerical result is a useful contrast between search score and selection performance. Even when optimization finds increasingly high candidate scores on the 1,000-image development subset, the best candidates are not uniformly better than the unchanged checkpoint on the 1,500-image selection subset. This is consistent with development adaptation or selection overfitting but does not isolate a causal mechanism without an additional prospective test. Moreover, the apparent monotonic optimization trend at $B=100$ and $250$ rests on two or one seeds, respectively.

\begin{figure*}[t]
\centering\includegraphics[width=.89\textwidth]{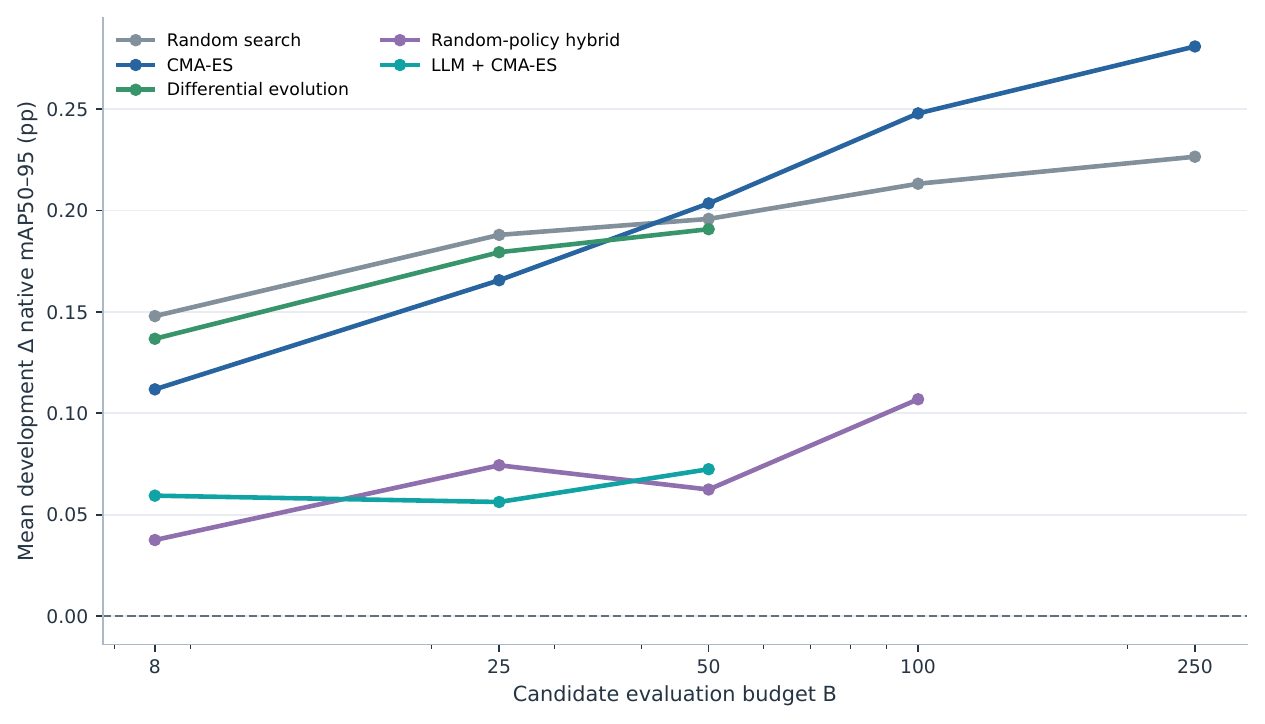}
\caption{Phase V mean development mAP50--95 changes versus candidate-evaluation budget, regenerated directly from archived per-seed records. Budget-specific sample sizes differ; the $B=250$ result contains one seed.}
\label{fig:budget}
\end{figure*}

\subsection{Budget-dependent LLM and hybrid performance}
The source matrix also records LLM-feedback, failure-aware, matched random-policy, and hybrid LLM--CMA-ES rows at selected budgets. Their inference overhead is substantial relative to numerical candidate search, although total time depends on implementation and hardware. At $B=8$, the real LLM-feedback method consumed 36,469 reported tokens over the experiment group and reached +0.02 development points versus -0.00 selection points; failure-aware LLM used 52,176 tokens and reached +0.02 and -0.01 points. At the same budget, the hybrid LLM--CMA-ES reached +0.06 development and -0.05 selection points, compared with +0.04 and -0.08 points for a random-policy hybrid. These are not paired significance results by themselves.

\begin{table}[t]
\caption{Selected Phase V LLM and random-policy hybrid rows from the packaged budget table (native mAP changes in percentage points).}
\label{tab:llmrows}
\centering\footnotesize
\begin{tabular}{llrr}
\toprule
$B$ & Policy & Dev $\Delta$ & Sel. $\Delta$\\
\midrule
8 & LLM feedback & +0.02 & $\approx$0.00\\
8 & Failure-aware LLM & +0.02 & -0.01\\
8 & Random-policy hybrid & +0.04 & -0.08\\
8 & LLM--CMA-ES hybrid & +0.06 & -0.05\\
25 & LLM feedback & +0.06 & -0.02\\
25 & Failure-aware LLM & +0.03 & +0.01\\
25 & Random-policy hybrid & +0.07 & -0.07\\
25 & LLM--CMA-ES hybrid & +0.06 & -0.01\\
50 & Random-policy hybrid & +0.06 & -0.01\\
50 & LLM--CMA-ES hybrid & +0.07 & -0.05\\
\bottomrule
\end{tabular}
\end{table}

\section{Results IV: Does Language-Model Guidance Add Value?}
\subsection{Matched paired statistical comparison}
The final paired statistical table compares the \emph{seed-level development changes} of the LLM hybrid with those of the random-policy hybrid under the same budgets. This is a more direct test of incremental LLM selection benefit than comparing the LLM against the unchanged detector. Table~\ref{tab:paired} preserves the reported paired difference, paired confidence interval and two-sided $p$ value.

\begin{table}[t]
\caption{Archived paired comparison of hybrid LLM--CMA-ES vs. matched random-policy hybrid. Differences are in native development mAP percentage points; $N$ denotes seed pairs.}
\label{tab:paired}
\centering\footnotesize
\begin{tabular}{rrrr}
\toprule
$B$ ($N$) & LLM$-$random & 95\% CI & $p$\\
\midrule
8 (5) & +0.02 & [-0.09,+0.13] & 0.612\\
25 (5) & -0.02 & [-0.07,+0.03] & 0.369\\
50 (3) & +0.01 & [-0.08,+0.10] & 0.694\\
\bottomrule
\end{tabular}
\end{table}

The intervals are wide relative to differences of a few hundredths of a percentage point, especially at $N=3$. These results do not establish that the methods are equal, that LLM reasoning has zero effect, or that improvements are impossible. They indicate that the evaluated LLM policy did not demonstrate superiority under the tested conditions. A future confirmatory comparison would preregister a practically relevant equivalence or superiority margin and preserve untouched test data.

\begin{figure}[t]
\centering\includegraphics[width=\columnwidth]{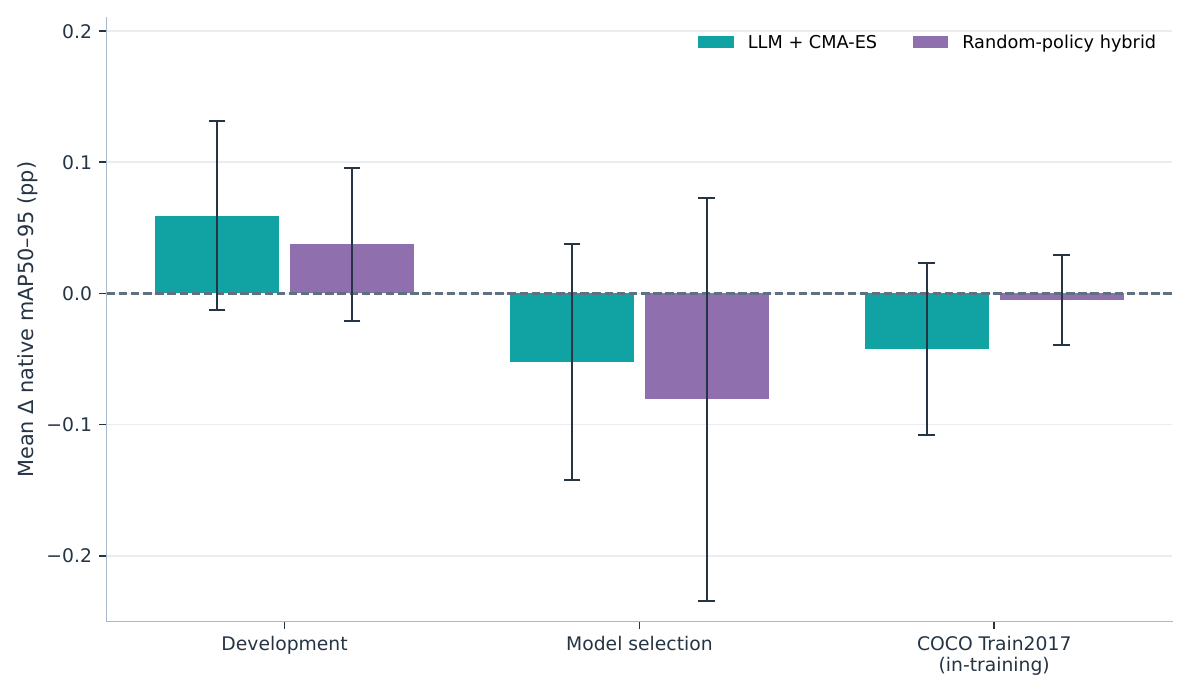}
\caption{Phase IV five-seed comparison of LLM-guided and randomly bounded hybrid CMA-ES on the development set, model-selection set, and an additional COCO Train2017 subset. Error bars are source-reported confidence intervals; the Train2017 subset is not an independent test of the pretrained detector.}
\label{fig:ablation}
\end{figure}

\subsection{Feedback, diagnostics and plausible failure mechanisms}
The Phase III ablation separated zero-shot LLM proposals, LLM with metric feedback, failure-aware LLM, and hybrid LLM--CMA-ES. Their development-set gains were small and some decreased selection accuracy. Detailed diagnostics can improve the specificity of natural-language hypotheses, but under limited data such specificity may target sample-dependent patterns rather than generalizable weight directions. Conversely, a constrained LLM subspace may avoid severe perturbations simply because its action norm is small. Both are explanations consistent with the observed numbers, not demonstrated mechanisms.

\begin{figure*}[t]
\centering\includegraphics[width=.89\textwidth]{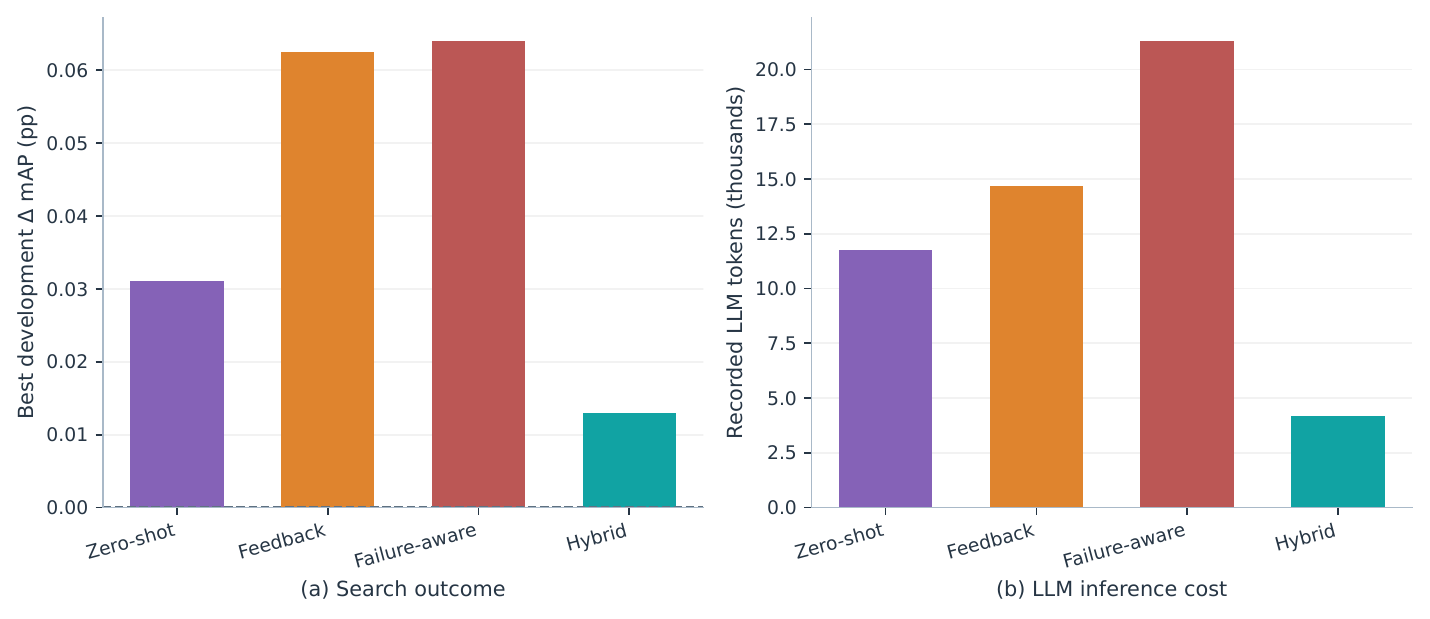}
\caption{Phase III LLM-guidance ablation using archived evaluation summaries. Left: best development mAP50--95 improvement; right: recorded token usage. This comparison is exploratory and does not establish that added feedback improves generalization.}
\label{fig:llmablate}
\end{figure*}

\subsection{Computational overhead}
A fair efficiency comparison must include LLM inference, candidate evaluation, retry overhead, and any postprocessing or parsing failures. The archive records tens of thousands of LLM tokens at several budgets, while random policies have no LLM token cost. Nevertheless, token totals in the source tables may represent all runs pooled at a budget, whereas wall times are means; these denominators should be harmonized before inferring per-candidate relative costs. The defensible conclusion is qualitative: real language-model interaction adds a measurable computational component not required by conventional numerical policies.

\begin{figure*}[t]
\centering\includegraphics[width=.89\textwidth]{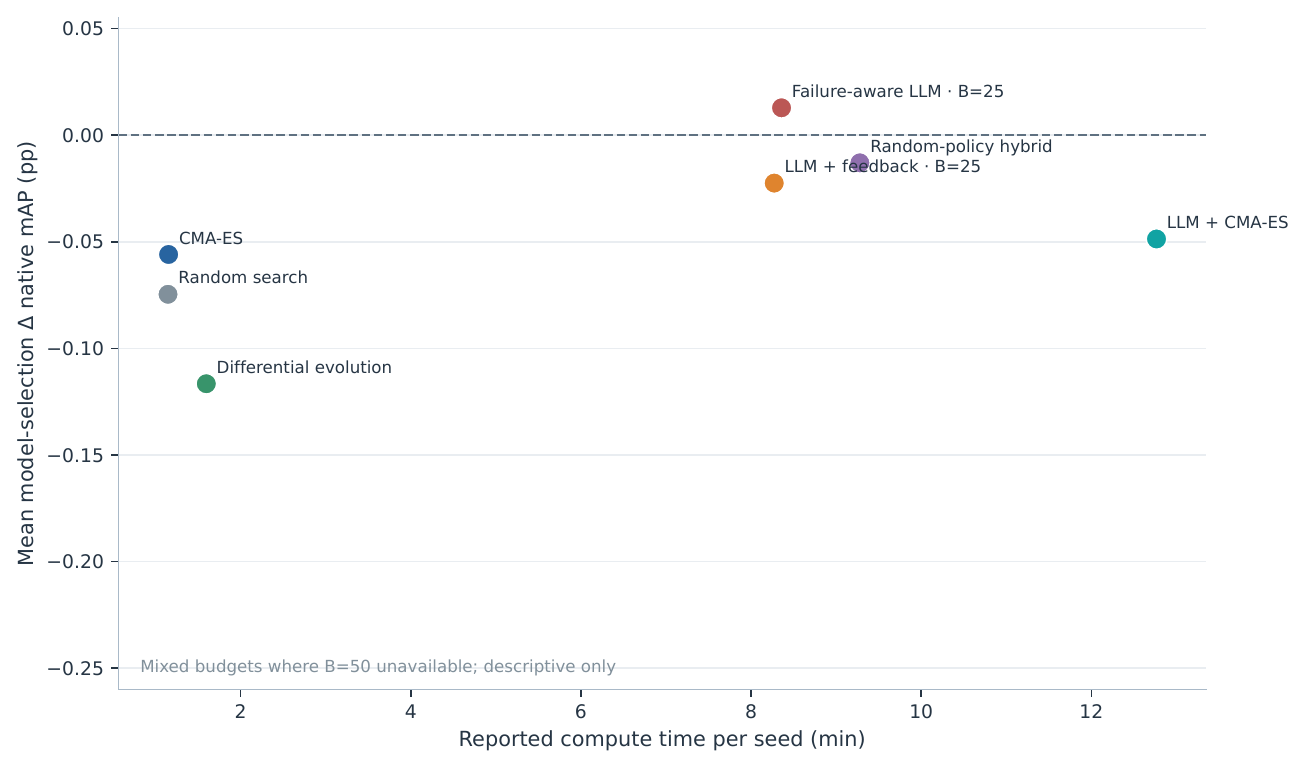}
\caption{Phase V descriptive comparison of reported compute time and model-selection performance. When budget $B=50$ records are unavailable, lower-budget records are shown and identified. Mixed budgets prohibit a formal Pareto-efficiency claim.}
\label{fig:cost}
\end{figure*}

\section{Results V: Model Capacity and YOLO26m Replication}
\subsection{Cross-model comparison}
The archive evaluates YOLO26n, YOLO26s and YOLO26m, with differing fused parameter counts and feature-block repetition. Single-seed cross-model CMA-ES results include an exploratory positive YOLO26m selection difference of approximately +0.04 points, while YOLO26n had a negative result. A transferred small-model editing policy was not reliably beneficial on larger models. Because these are short-budget and mostly single-seed results, they do not establish a capacity law or prove that a particular architecture is inherently more tolerant of perturbations.

\subsection{Five-seed YOLO26m replication}
To check the initially positive YOLO26m result, the source project ran CMA-ES in a five-stage subspace with $B=15$ evaluations using seeds 42--46. Seed 42 had been examined in earlier exploratory work, while seeds 43--46 were added subsequently. Table~\ref{tab:yolom} reports native and official COCO API differences. The mean native selection change was approximately +0.001 points, and the mean official COCO AP change was +0.004 points with a reported 95\% interval $[-0.026,+0.034]$. Thus the replicated mean is near zero, even though individual seeds have positive or negative differences.

\begin{table}[t]
\caption{Targeted YOLO26m five-seed replication on 1,500 selection images. All numbers are percentage-point changes versus the same untouched baseline.}
\label{tab:yolom}
\centering\footnotesize
\begin{tabular}{rrr}
\toprule
Seed & Native $\Delta$mAP & COCO API $\Delta$AP\\
\midrule
42 (exploratory) & +0.043 & +0.035\\
43 & -0.051 & -0.030\\
44 & +0.014 & +0.015\\
45 & +0.012 & +0.005\\
46 & -0.016 & -0.006\\
\midrule
Mean & +0.001 & +0.004\\
95\% CI & [-0.044,+0.045] & [-0.026,+0.034]\\
\bottomrule
\end{tabular}
\end{table}

\begin{figure*}[t]
\centering\includegraphics[width=.89\textwidth]{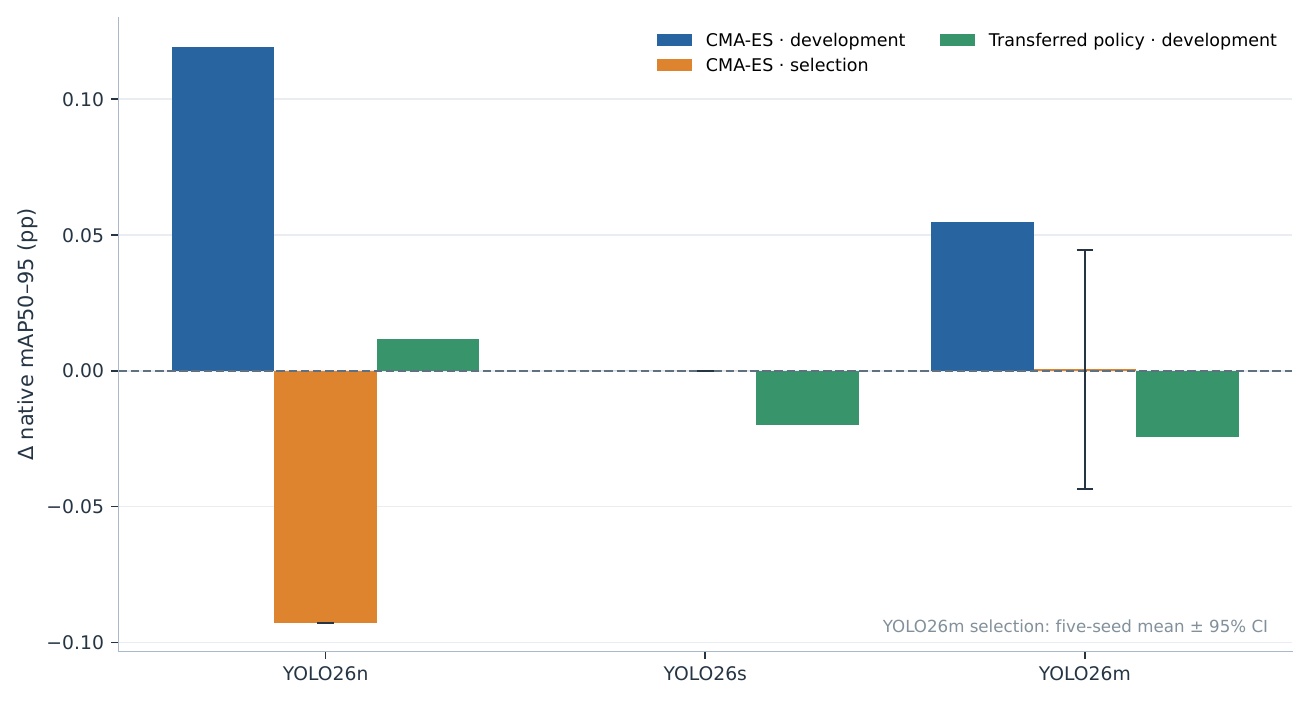}
\caption{Archived YOLO26n/s/m cross-model editing comparisons, including the YOLO26m five-seed replication confidence interval for selection native mAP. Single-seed estimates for other model scales remain exploratory.}
\label{fig:capacity}
\end{figure*}

\section{Reproducibility and Forensic Audit of LLM Actions}
\subsection{What was verified}
The software package includes immutable-checkpoint checks, gradient checks, disjointness checks for specified manifests, and action validation tests. The reported tests pass under the archived execution environment. The final report documents the use of a real local language model with response tokens and latency captured from Ollama rather than synthetic counters, thereby addressing the original Gaussian-simulator mislabeling.

\subsection{Retrospective versus contemporaneous validity}
Table~\ref{tab:audit} gives the final strict-schema audit. Rejection by the 2026 finalized validator does not imply that earlier proposals were rejected online by the then-current code. More importantly, without complete historic intermediate checkpoint hashes, one cannot prove which tensor states were used to evaluate every early proposal. This prevents strong claims about historical fallback counts. Such caveats are substantive evidence limitations, not merely a formatting issue.

\begin{table}[t]
\caption{Final strict-schema retrospective proposal audit. These counts do not establish historical online execution outcomes.}
\label{tab:audit}
\centering\footnotesize
\begin{tabular}{lrr}
\toprule
Retrospective category & Count & Share\\
\midrule
Archived parsed proposals & 225 & 100\%\\
Schema-valid now & 53 & 23.6\%\\
Schema-invalid now & 172 & 76.4\%\\
Verified complete original tensor trail & N/A & N/A\\
\bottomrule
\end{tabular}
\end{table}

\subsection{Minimum reproducibility specification for future studies}
Future work should save each exact prompt and unmodified response, the schema version, the proposed and accepted edit specification, all targeted tensor identifiers, a hash of the resulting checkpoint, the evaluation image-ID manifest, inference configuration, per-image predictions, and the software environment. Any invalid action, retry or fallback must have an explicit event label. The numerical optimizer's parameter bounds and candidate-evaluation budget should be serialized. Such an artifact would allow later investigators to distinguish whether a human-readable LLM hypothesis actually caused the evaluated parameter change.

\section{Discussion}
\subsection{Why development-set gains need not generalize}
Selecting the best among many candidate scores on a finite dataset creates a multiple-comparisons problem even if each candidate is produced without gradients. A small change in detection AP can reflect the interaction of image composition, object size, confidence sorting and postprocessing, rather than a general improvement in the detector. The observation that larger budgets produce better development scores but often worse selection scores is compatible with adaptive validation-set fitting~\cite{dwork2015,recht2019}. It does not by itself prove a specific loss-landscape mechanism.

\subsection{What is the LLM contribution?}
The source project demonstrates that an LLM can provide syntactically interpretable proposals and high-level subspace choices. However, interpretability of a \emph{proposal} is different from predictive utility of the proposal. The matched random-policy hybrid provides a direct empirical check: when random choices share the same numerical optimizer and broad action constraints, the observed LLM-minus-random differences are small and uncertain. As a result, the central claim should be \emph{no demonstrated incremental advantage} rather than proof that language models cannot optimize weights.

\subsection{Relationship to model merging and parameter-efficient adaptation}
Model soups, task arithmetic, and TIES merging operate on trained or fine-tuned weight states containing task information~\cite{wortsman2022,ilharco2023,yadav2023}; LoRA introduces trainable low-rank factors~\cite{hu2022}. The WSP intervention studied here is different: it searches a constrained family of modifications of a fixed pretrained detector using evaluation scores and no gradients. Consequently, prior successful model averaging or trained low-rank adaptation does not imply that these gradient-free interventions should achieve comparable gains.

\subsection{Scientific value of null or near-null results}
A benchmark showing that visually plausible LLM proposals do not beat a well-matched random policy can help prevent overclaiming, particularly when model-generated rationales appear causally convincing. It also identifies a reproducibility bottleneck: the semantic decision, the accepted action and the actual resulting tensor state must be linked. The result can motivate future methods with validated causal features, stronger priors, or richer action parameterizations, but those possibilities remain hypotheses.

\subsection{Limitations and threats to validity}
\emph{Evaluation provenance:} The \texttt{fresh} partition has contradictory Train2017/Val2017 lineage in the archived documents; it is not used here for independent generalization claims. \emph{Selection reuse:} The model-selection partition was inspected across research phases and therefore is not pristine. \emph{Mixed metrics:} Native and COCO API AP differ and must not be blended. \emph{Source discrepancy:} The Phase V narrative and its packaged tabular series disagree on some multi-budget outcomes; this manuscript explicitly selects the packaged table and flags the conflict. \emph{Limited replication:} Budgets of 100 and 250 have only two and one seed; the YOLO26m replication retains an earlier exploratory seed. \emph{Action fidelity:} A retrospective schema check cannot prove historic tensor execution. \emph{Mechanistic uncertainty:} Observed sensitivity is not sufficient to infer Hessian sharpness, systematic confidence recalibration, or layer-specific causal effects. \emph{Generalization scope:} Results apply to particular pretrained COCO checkpoint scales, bounded edit types, and tested settings, not every possible weight-space policy.

\section{Recommendations for a Confirmatory Follow-Up}
A future confirmatory experiment should preregister a single policy specification, freeze the architecture-specific layer map and action bounds, and report the complete candidate evaluation budget including retries. It should separate the dataset used for search, any model-selection dataset, and an untouched evaluation dataset whose image identifiers and pretraining history have been verified. Results should be reported across enough seeds to estimate the optimizer-level variation, and image-level paired uncertainty should be computed for each finalized candidate. A meaningful practical improvement threshold should be defined before inspecting the final data. For LLM comparisons, the same number of \emph{valid} evaluated proposals and the same numerical downstream budget should be counted; language-model tokens and latency must be reported in addition to GPU time.

\section{Conclusion}
This five-phase study documents direct gradient-free edits of pretrained YOLO26 models using deterministic interventions, random and evolutionary search, genuine local-LLM proposals, failure diagnostics, and hybrid LLM--CMA-ES methods. Low-dimensional search produces small development-set gains, but the archived selection comparisons do not establish consistent improvement. The paired hybrid LLM-versus-random tests fail to demonstrate an incremental LLM advantage at budgets of 8, 25, and 50, and a targeted five-seed YOLO26m replication gives a near-zero mean official COCO AP difference. The strongest contribution is therefore an audited empirical characterization of direct weight editing and its evaluation risks, not an accuracy-enhancing detector. Future claims should rest on frozen policies, genuinely untouched test data, and end-to-end proposal-to-tensor evidence.

\section*{Data and Code Availability}
The accompanying project archive contains implementation modules, configurations, figure assets, selected raw results, derived tables and reproducibility scripts. Original large pretrained checkpoints and the COCO images are external and identified in the archived documentation.

\appendices
\section{Detailed Phase III and IV Method Matrix}
The preliminary seven-method Phase III experiment used a single 15-evaluation budget, and its results should not be treated as definitive estimates. Table~\ref{tab:phase3full} provides the source values for all reported policies. In this table, the development and selection scores use the native evaluator. The source also reports a 1,000-image ``fresh'' series; we omit that column from numerical inference because its provenance is unresolved. The relative ranking of methods is budget- and cohort-specific.
\begin{table*}[t]
\caption{Complete Phase III seven-method comparison at budget $B=15$ (source: Phase III archived Table 12). Model baseline mAP: 42.08\% development and 39.17\% selection. Values in parentheses are absolute percentage-point changes from the unchanged model.}
\label{tab:phase3full}
\centering\footnotesize
\begin{tabular}{lrrlll}
\toprule
Optimizer & LLM calls & Prompt / completion tokens & LLM time (s) & Development native mAP & Selection native mAP\\
\midrule
Random search & 0 & 0 / 0 & 0 & 42.19 (+0.11) & 38.98 (-0.19)\\
Differential evolution & 0 & 0 / 0 & 0 & 42.21 (+0.13) & 39.09 (-0.08)\\
CMA-ES & 0 & 0 / 0 & 0 & 42.20 (+0.12) & 39.08 (-0.09)\\
Real LLM (zero-shot) & 14 & 7,440 / 4,294 & 133.8 & 42.11 (+0.03) & 39.14 (-0.03)\\
Real LLM + feedback & 14 & 9,848 / 4,832 & 140.4 & 42.14 (+0.06) & 39.17 (+0.00)\\
Failure-aware LLM & 14 & 15,059 / 6,222 & 172.9 & 42.14 (+0.06) & 38.99 (-0.18)\\
Hybrid LLM--CMA-ES & 4 & 2,959 / 1,209 & 56.1 & 42.09 (+0.01) & 39.16 (-0.01)\\
\bottomrule
\end{tabular}
\end{table*}

The Phase IV $B=8$ comparison adds five seeds for each policy and is shown in Table~\ref{tab:phase4full}. It should not be merged numerically with the single-seed Phase III table because the search policy, action validator, seed set, and available information changed. Notably, the Phase IV source table itself labels the metric as official COCO AP, whereas the experiment runner uses the native objective. We adopt the implementation-consistent ``native mAP'' label for these entries and disclose the inconsistency.

\begin{table*}[t]
\caption{Complete Phase IV multi-seed comparison, $B=8$, five seeds. Values are native development and selection mAP changes in percentage points, with development SD from the source table. All $N=5$.}
\label{tab:phase4full}
\centering\small
\begin{tabular}{lrrrl}
\toprule
Optimization method & Dev mean $\Delta$ & Dev SD & Selection mean $\Delta$ & Total recorded LLM tokens\\
\midrule
Random search & +0.15 & 0.06 & -0.11 & 0\\
CMA-ES & +0.11 & 0.06 & -0.03 & 0\\
Differential evolution & +0.14 & 0.08 & -0.05 & 0\\
LLM with metric feedback & +0.02 & 0.02 & $\approx$0.00 & 36,469\\
Failure-aware LLM (strict) & +0.02 & 0.01 & -0.01 & 52,176\\
LLM--CMA-ES hybrid & +0.06 & 0.06 & -0.05 & 12,246\\
Random-policy hybrid & +0.04 & 0.05 & -0.08 & 0\\
\bottomrule
\end{tabular}
\end{table*}

\section{A More Granular View of Phase V Budget Scaling}
Averaged development and selection statistics obscure within-seed variation and differing cost denominators. The Phase V archive has five seeds at budgets 8 and 25 for numerical methods, while the two LLM-only methods have three seeds at budget 25. At budget 50, three seeds are reported for random, DE, CMA-ES, and both hybrid controls; at budget 100, only the numerical methods and random-policy hybrid are represented; and at 250, numerical methods have only one seed. Thus, the term ``multi-budget'' describes a heterogeneous experimental matrix, not a balanced factorial design. Comparisons should be restricted to shared budgets and similarly defined cohorts.

\begin{table*}[t]
\caption{Additional LLM-related Phase V rows and cost records from packaged Table 17. Development/selection values are native mAP changes (pp); mean time and token counts retain their original source denominators.}
\label{tab:llmexpanded}
\centering\small
\begin{tabular}{rrlrrrr}
\toprule
$B$ & $N$ & Policy & Dev $\Delta$ & Selection $\Delta$ & Mean time (s) & Total tokens\\
\midrule
8 & 5 & LLM feedback & +0.02 & $\approx$0.00 & 160.6 & 36,469\\
8 & 5 & Failure-aware LLM & +0.02 & -0.01 & 157.0 & 52,176\\
8 & 5 & Random-policy hybrid & +0.04 & -0.08 & 110.3 & 0\\
8 & 5 & LLM--CMA-ES hybrid & +0.06 & -0.05 & 144.4 & 12,246\\
25 & 3 & LLM feedback & +0.06 & -0.02 & 496.3 & 77,721\\
25 & 3 & Failure-aware LLM & +0.03 & +0.01 & 501.5 & 116,364\\
25 & 5 & Random-policy hybrid & +0.07 & -0.07 & 273.8 & 0\\
25 & 5 & LLM--CMA-ES hybrid & +0.06 & -0.01 & 371.8 & 35,436\\
50 & 3 & Random-policy hybrid & +0.06 & -0.01 & 556.7 & 0\\
50 & 3 & LLM--CMA-ES hybrid & +0.07 & -0.05 & 765.9 & 45,718\\
100 & 2 & Random-policy hybrid & +0.11 & -0.02 & 1119.3 & 0\\
\bottomrule
\end{tabular}
\end{table*}

The critical question is whether additional candidate evaluations yield \emph{selection improvement}, not whether additional candidates can be found with higher development fitness. Figure~\ref{fig:budget} depicts the trajectory and Table~\ref{tab:budgets} records the numerical values. Because selecting the maximum observed score is an adaptive operation, even an unbiased candidate-generation mechanism can induce an optimistic best-of-many estimate~\cite{bergstra2012,dwork2015}. This observation is particularly salient when numerical gains are well below one percentage point.

\section{Interpretation of Diagnostic Figures and Failure Modes}
\begin{figure*}[t]
\centering\includegraphics[width=.9\textwidth]{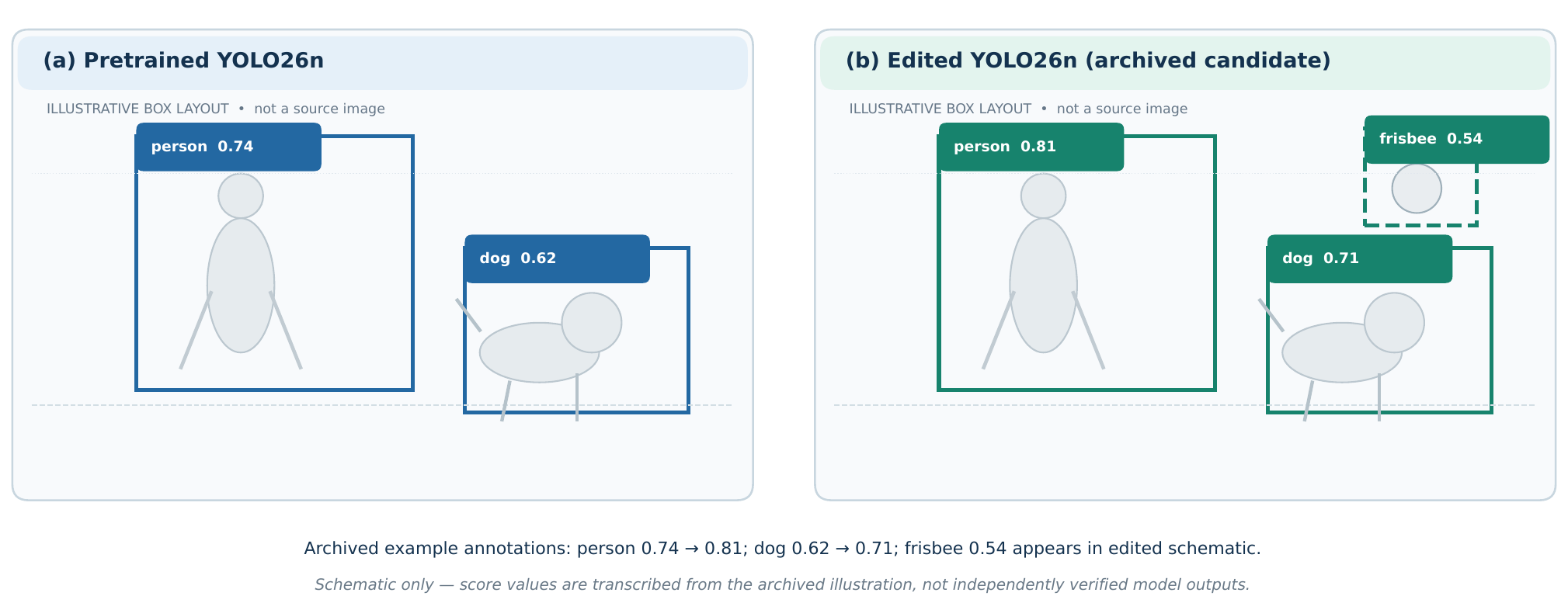}
\caption{Redesigned schematic of the archived original-versus-edited YOLO26 detection example. The displayed object labels and confidence scores are transcribed from the original schematic and are not independently verified COCO predictions. No original scene photograph or new inference results are available for this comparison.}
\label{fig:qualitative}
\end{figure*}

The archived visual materials contain object-size comparisons, per-class AP changes, sensitivity heatmaps, score distributions, schematic detection boxes, and cost--accuracy summaries. They are retained as experimental illustrations rather than as independently recalculated quantities. In particular, Fig.~\ref{fig:qualitative} is a redrawn archival schematic, not a verified example image: its confidence annotations must not be interpreted as an independent detection benchmark. Qualitative examples should never replace AP statistics, because an appealing improvement on a selected example may coexist with worse ranking across the whole dataset.

The classification-head and regression-head ablations suggest that detection quality depends on which branch is edited. However, a confidence distribution change can affect measured precision and recall without necessarily reflecting improved feature representations. Similarly, magnitude pruning of one percent with near-zero AP change establishes only that the specific metric is insensitive to that sparsification amount in the tested configuration; it does not establish actual throughput, energy, or memory savings. Future work could evaluate hardware-level effects independently, using standardized deployment and corruption protocols~\cite{hendrycks2019,goodfellow2015}.

\section{Verification Checklist for an IEEE Submission}
Before external submission, the following documentary checks should be completed by the authors: (i) reconcile the \texttt{fresh\_eval} image manifest with COCO Train2017 and Val2017 identifiers; (ii) reconcile Phase V narrative values with the archived Table 17 series and regenerate any figures that rely on stale aggregations; (iii) identify the checkpoint format used in each count, distinguishing fused from unfused parameters; (iv) preserve checkpoint and image-ID hashes; (v) attach exact accepted LLM proposals to evaluated tensor hashes wherever data exist; (vi) add author names, ORCID identifiers, affiliations and disclosures; and (vii) deposit a versioned code and result package with a persistent repository identifier. None of these checks requires asserting additional model training or fabricating missing evaluation results.

\end{document}